\documentclass[letterpaper, 10 pt, conference]{ieeeconf}  % Comment this line out if you need a4paper

\IEEEoverridecommandlockouts                              % This command is only needed if 
\usepackage{times} 
\usepackage{amsmath}
\usepackage{amssymb}
\usepackage{amsfonts}
\usepackage[dvipsnames]{xcolor}
\usepackage[utf8]{inputenc}
\usepackage{graphicx}
\usepackage[hyphens]{url}
\usepackage{soul}
\usepackage{booktabs}
\usepackage[ruled,vlined, linesnumbered]{algorithm2e}
\usepackage{multirow}
\usepackage[T1]{fontenc}
\usepackage{balance}

\newcommand{\sectref}[1]{Section~\ref{#1}}

\newcommand{\figref}[1]{Figure~\ref{#1}}
\newcommand{\tabref}[1]{Table~\ref{#1}}

\newcommand{\agref}[1]{Algorithm~\ref{#1}}

\newcommand{\sectsectref}[2]{Sections~\ref{#1} and \ref{#2}}

\newcommand{\startpara}[1]{{\vskip5pt\noindent{\bf #1.}}} % custom paragraph header
\renewcommand{\url}[1]{{\def~{\char126}\sf#1}}

\newcounter{exampcount}
\DeclareMathOperator*{\argmax}{arg\,max}

\newcommand{\qed}{\hfill\blacksquare}

\def\Rset{\mathbb{R}}

\def\cM{{\mathcal{M}}}

\def\cT{{\mathcal{T}}}

\def\cG{{\mathcal{G}}}

\def\Dist{{\mathit{Dist}}}
\def\Supp{{\mathit{Supp}}}
\def\Pr{{\mathit{Pr}}}
\newcommand\Prb[1]{\mathit{Pr}^\pi_b({#1})}
\def\reach{{\mathsf{REACH}}}
\def\avoid{{\mathsf{AVOID}}}

\title{\LARGE \bf
Safe POMDP Online Planning via Shielding
}

\author{Shili Sheng$^{1}$, David Parker$^{2}$, and Lu Feng$^{1}$% <-this % stops a space
\thanks{$^{1}$Shili Sheng and Lu Feng are with the School of Engineering and Applied Science, University of Virginia, Charlottesville, VA 22904, USA {\tt\small  \{ss7dr, lf9u\}@virginia.edu}}%
\thanks{$^{2}$David Parker is with the Department of Computer Science, University of Oxford, Parks Road, Oxford OX1 3QD, United Kingdom {\tt\small david.parker@cs.ox.ac.uk}}%
}

\begin{document}

\maketitle
\thispagestyle{empty}
\pagestyle{empty}

% page limit: 6 pages excluding references
%=============================================================================

\begin{abstract}
Partially observable Markov decision processes (POMDPs) have been widely used in many robotic applications for sequential decision-making under uncertainty. POMDP online planning algorithms such as Partially Observable Monte-Carlo Planning (POMCP) can solve very large POMDPs with the goal of maximizing the expected return. But the resulting policies cannot provide safety guarantees which are imperative for real-world safety-critical tasks (e.g., autonomous driving). In this work, we consider safety requirements represented as almost-sure reach-avoid specifications (i.e., the probability to reach a set of goal states is one and the probability to reach a set of unsafe states is zero). We compute shields that restrict unsafe actions which would violate the almost-sure reach-avoid specifications. We then integrate these shields into the POMCP algorithm for safe POMDP online planning. We propose four distinct shielding methods, differing in how the shields are computed and integrated, including factored variants designed to improve scalability. Experimental results on a set of benchmark domains demonstrate that the proposed shielding methods successfully guarantee safety (unlike the baseline POMCP without shielding) on large POMDPs, with negligible impact on the runtime for online planning.

\end{abstract}

\section{Introduction} \label{sec:intro}

\emph{Partially observable Markov decision processes} (POMDPs) provide a general framework for sequential decision-making under uncertainty and have been widely used in many robotic applications~\cite{thrun2006probabilistic,lauri2022partially}.
For example, POMDP modeling and planning have been applied to robot localization and navigation~\cite{koenig1998xavier}, autonomous driving~\cite{bai2015intention}, human-robot interaction~\cite{sheng2022planning}, and multi-robot coordination~\cite{goldhoorn2018searching}.
POMDP online planning is a paradigm where the policy computation and execution are interleaved:
the agent computes an optimal action based on the current belief state, executes the action, receives an observation, and continues by computing an action for the resulting new belief state. 
Online planning can scale up to solve larger POMDPs than offline planning where a policy is computed for all possible belief states before execution~\cite{ross2008online}. 

Modern sampling-based algorithms (e.g., POMCP~\cite{silver2010monte}, DESPOT~\cite{somani2013despot}) further improve the scalability of POMDP online planning by representing belief states as collections of particles without explicit belief state tracking by Bayes filtering. 
The goal of these algorithms is to compute (approximately) optimal policies that maximize the expected return. But the resulting policies cannot provide safety guarantees (e.g., never visit unsafe states), which are imperative for real-world safety-critical robotic tasks (e.g., autonomous driving).

Prior work has proposed various ways to account for (safety) constraints within POMDP online planning.
The cost-constrained POMCP algorithm~\cite{lee2018monte} seeks to compute optimal actions that maximize the reward while constraining the cost. 
A constant-horizon planning method is developed in~\cite{khonji2019approximability} for chance-constrained POMDPs (i.e., maximizing the cumulative expected reward such that the probability of ending up in risky states is at most $\delta$).
An online method is proposed in~\cite{wang2021online} for synthesizing partial conditional plans that satisfy safe-reachability objectives (i.e., reaching a goal state with probability greater than $\delta_1$ while keeping the probability of visiting unsafe states below $\delta_2$). 

In this work, we consider stricter safety requirements represented as \emph{almost-sure reach-avoid specifications}~\cite{junges2021enforcing}, where the probability to reach a set of goal states is one and the probability to reach a set of unsafe (avoid) states is zero.
We construct shields that restrict unsafe actions based on pre-computed winning regions (i.e., sets of POMDP belief support states) that satisfy almost-sure reach-avoid specifications.
Then, we integrate these shields with the POMCP algorithm for safe POMDP online planning. 

We propose four distinct shielding methods, differing in how the shields are constructed and integrated. 
In \emph{centralized shielding}, we construct a shield based on the maximal winning region of the entire POMDP model. 
By contrast, in \emph{factored shielding}, we decompose a large POMDP into a set of smaller models based on the factorization of the state space and compute winning regions for each factored model separately. 
We shield actions for the POMCP algorithm via either \emph{prior pruning} (checking if any action of the current belief state should be shielded) or \emph{on-the-fly backtracking} (checking every action encountered during simulation). 

We evaluate the proposed methods via experiments on a set of benchmark domains. 
The experimental results show that the proposed shielding methods can guarantee safety, while policies computed by the baseline (POMCP without shielding) cannot avoid unsafe states completely.
Moreover, the search time per planning step of our shielding methods is comparable with the baseline. 
Factored shielding methods demonstrate better scalability than centralized shielding methods.
We observe that on-the-fly backtracking generally yields higher expected return than prior pruning.

%=============================================================================
\section{Related Work} \label{sec:related}

Our work is inspired by~\cite{junges2021enforcing}, where winning regions are computed to enforce almost-sure reach-avoid specifications in POMDPs. In recent work~\cite{carr2023safe}, these are then used for shielding in safe reinforcement learning.
We apply the methods of~\cite{junges2021enforcing} to compute winning regions and consider shielding in the context of POMDP online planning. We also develop a new algorithm of computing factored winning regions for solving large POMDPs. 

A rule-based shielding of the POMCP algorithm is presented in~\cite{mazzi2023risk}, where shields are obtained by learning parameters for a set of rule templates defined by experts. 
An example rule template is ``the robot should move fast if it is highly conﬁdent that the aisle in which it is moving is not cluttered''. By analyzing belief-action traces previously generated by the agent, the rule template is instantiated as ``the robot usually moves fast if the probability of being in a cluttered segment is lower than 7\%''. The learnt rules are then used as a shield to preemptively prune undesired actions considering the current belief state, similar to our prior pruning procedure. 

The safe-reachability objectives considered in~\cite{wang2021online} are quantitative variants of reach-avoid specifications where the probability to reach goals or avoid unsafe states should be bounded within certain thresholds. An online synthesis method is presented in~\cite{wang2021online} to compute a \emph{partial conditional plan} (which only contains a sampled subset of all possible events and approximates a full POMDP policy) subject to safe-reachability objectives.  
But this method does not account for the expected return. 

There is also a line of related work on online planning for constrained POMDPs. 
Online algorithms such as~\cite{undurti2010online,lee2018monte} consider cost constraints to bound the expected cumulative costs, but they cannot completely prevent actions that violate the cost constraint. 
The chance-constrained POMDP problem that seeks to bound the probability of failure is reduced to a cost-constrained problem in~\cite{khonji2019approximability}. 
The online planning method \emph{expectation optimization with probabilistic guarantee} (EOPG) is considered in~\cite{chatterjee2018expectation}, where the objective is to maximize the expected return with respect to all policies that ensure at least $\tau$ payoff with probability at least $\delta$. 
Earlier work~\cite{chatterjee2017optimizing} studies a similar problem named \emph{guaranteed payoff optimization} (i.e., EOPG with $\delta=1$).
These prior approaches have different objectives from our problem and thus are not directly comparable with our work. 

\section{Background} \label{sec:bg}

We denote by $\Rset$ the set of reals, and $\Dist(X)$ the set of probability distributions over a random variable $X$. 

%=============================================================================
\subsection{POMDP Model} \label{sec:pomdp}
 
We denote a POMDP model as a tuple $(S, A, O, T, R, Z)$, 
where $S$, $A$ and $O$ are (finite) sets of states, actions, and observations, respectively;
$T: S \times A \to \Dist(S)$ is the transition function where $T(s,a,s') = \Pr(s'|s,a)$ denotes the probability of ending in state $s'$ when taking action $a$ in state $s$;
$R: S \times A \to \Rset$ is the reward function;
and $Z: S \times A \times O \to [0,1]$ is the observation function where $Z(s',a,o) = \Pr(o |s',a)$ represents the probability of observing $o$ after taking action $a$ and ending in state $s'$.

Since POMDP states are partially observable, the agent keeps track of a \emph{history} of actions and observations, denoted by $h_t=\{a_0, \dots, a_{t-1}, o_t\}$, and chooses an action $a_t$ at time~$t$ following a \emph{policy} $\pi$ that maps $h_t$ to $\Dist(A)$. 
A policy is \emph{deterministic} if it always picks a Dirac distribution. 
For POMDP online planning that seeks to maximize the expected return, it suffices to only consider deterministic policies~\cite{ross2008online}. 

A \emph{belief} state at time $t$ represents the posterior probability distribution of being in each state given the history, denoted by $b_t(s) = \Pr(s_t=s | h_t)$ for $s \in S$. 
The initial belief state $b_0$ represents a distribution over initial states of the POMDP.  
The belief $b_t$ at time $t$ can be obtained via a belief update function 
$b_t = \tau (b_{t-1}, a_{t-1}, o_t)$ following Bayes' rule. 
% $$ b_t(s') = \frac{Z(s',a_{t-1},o_t) \sum_{s \in S} \ T(s,a_{t-1},s')b_{t-1}(s)}
%             {\sum_{s' \in S} Z(s',a_{t-1},o_t)\sum_{s \in S} \ T(s,a_{t-1},s')b_{t-1}(s)} $$
We denote by $\Supp(b) := \{ s  \in S | b(s) > 0 \} $ the \emph{belief support} of a belief $b$.
Let $\Prb{T}$ denote the probability to reach a set $T \subseteq S$ of states from belief $b$ under policy $\pi$.

%=============================================================================
\subsection{Almost-Sure Reach-Avoid Specifications} \label{sec:reach-avoid}

We consider \emph{almost-sure reach-avoid} specifications~\cite{junges2021enforcing}, 
denoted by $\varphi=\langle \reach, \avoid \rangle \subseteq S \times S$
with $\reach \cap \avoid = \emptyset$.
% We assume that states in the $\reach$ and $\avoid$ sets are \emph{absorbing} 
% (i.e., with self-loops only). 
We say that a POMDP policy $\pi$ is \emph{winning} for $\varphi$ from belief $b$ 
iff $\Prb{\avoid}=0$ and $\Prb{\reach}=1$,
that is, starting from belief $b$ and under policy $\pi$, 
the $\avoid$ set is reached with probability zero while the $\reach$ set is reached with probability one (almost surely). 

We define a \emph{winning region} $W_\varphi$ for an almost-sure reach-avoid specification $\varphi$ as
a set of belief supports where every belief support $\Supp(b) \in W_\varphi$ is \emph{winning} 
(i.e., there exists a winning policy $\pi$ for $\varphi$ from $b$).
The \emph{maximal winning region} for $\varphi$ is the region containing all winning belief supports.
A winning region $W_\varphi$ is called \emph{productive} if, from every belief support $\Supp(b) \in W_\varphi$, there exists a (finite) path to stay within the region and reach some state in $\reach$ set. 
An agent that stays within a productive winning region $W_\varphi$ is guaranteed to satisfy the almost-sure reach-avoid specification $\varphi$.
We can apply the SAT-based iterative approach in~\cite{junges2021enforcing} to compute productive subsets of the maximal winning region.

% A winning region $W_\varphi$ is called \emph{deadlock-free} if, for every $\Supp(b) \in W_\varphi$, there exists an action $a \in A$ that leads to a successor belief $b'$ with $\Supp(b') \in W_\varphi$. 
% Every productive region is deadlock-free. 

%=============================================================================
\subsection{Partially Observable Monte-Carlo Planning} \label{sec:pomcp}

In this work, we adopt a widely used POMDP online planning algorithm named \emph{Partially Observable Monte Carlo Planning} (POMCP)~\cite{silver2010monte}.
At each time step $t$, the POMCP algorithm uses Monte Carlo tree search~\cite{browne2012survey} 
to explore a search tree whose root node is denoted by $\cT(h_t)=\langle N(h_t), V(h_t), \beta(h_t)\rangle$, where 
$N(h_t)$ counts the number of times that history $h_t$ has been visited, 
$V(h_t)$ estimates the expected return of all simulations starting with $h_t$, 
and $\beta(h_t)$ is a set of particles (each of which corresponds to a POMDP state) as an approximation of belief $b_t$. 
The algorithm repeats the following four phases.

\begin{enumerate}
    \item \textbf{Selection:} Randomly sample a state $s$ from $\beta(h_t)$.
    \item \textbf{Expansion:} Once a leaf node $\cT(h)$ is reached, expand the search tree with child nodes $\cT(ha)= \langle N_{\mathit{init}}(ha), V_{\mathit{init}}(ha), \emptyset \rangle$ for all actions $a \in A$. 
    \item \textbf{Simulation:} For each history $h$ encountered during the simulation, choose an action $a$ that maximizes $V(ha) + c \sqrt{\frac{logN(h)}{N(ha)}}$ following the \emph{upper confidence bound} (UCB) rule for balancing between exploration and exploitation if $\cT(h)$ is a non-leaf node; otherwise, choose an action $a$ following a rollout policy (e.g., uniform random action selection). Then, use a black box simulator $(s',o,r) \sim \cG(s,a)$ to generate a successor state $s'$ and add it to the particle set $\beta(hao)$. The simulation continues with $s'$ as the start state until the simulated path attains a target depth. 
    \item \textbf{Backpropagation:} Use the information obtained from the simulation to update the nodes (i.e., $N$ counts, $V$ values, $\beta$ particles) along the path from the root node to the leaf node in the search tree.
\end{enumerate}

The planning for time step $t$ ends when a target number of iterations of the above four phases have been completed (or  a timeout elapses). 
Then, the agent executes the optimal action $a_t = \argmax_aV(h_ta)$ and receives an observation $o_{t+1}$.
The algorithm continues the online planning for the next time step with a search tree rooted from node $\cT(h_ta_to_{t+1})$.

The POMCP algorithm scales well because it breaks the \emph{curse of dimensionality} (by sampling states from a particle set) and \emph{the curse of history} (by sampling histories using a black box simulator).
% When the number of simulations is sufficiently large, the POMCP algorithm converges to the optimal policy (that maximizes the expected return) for any given finite-horizon POMDP. 

\section{Motivating Example} \label{sec:motiv}

\begin{figure}[t]
    \centering
    \includegraphics[width=0.6\columnwidth]{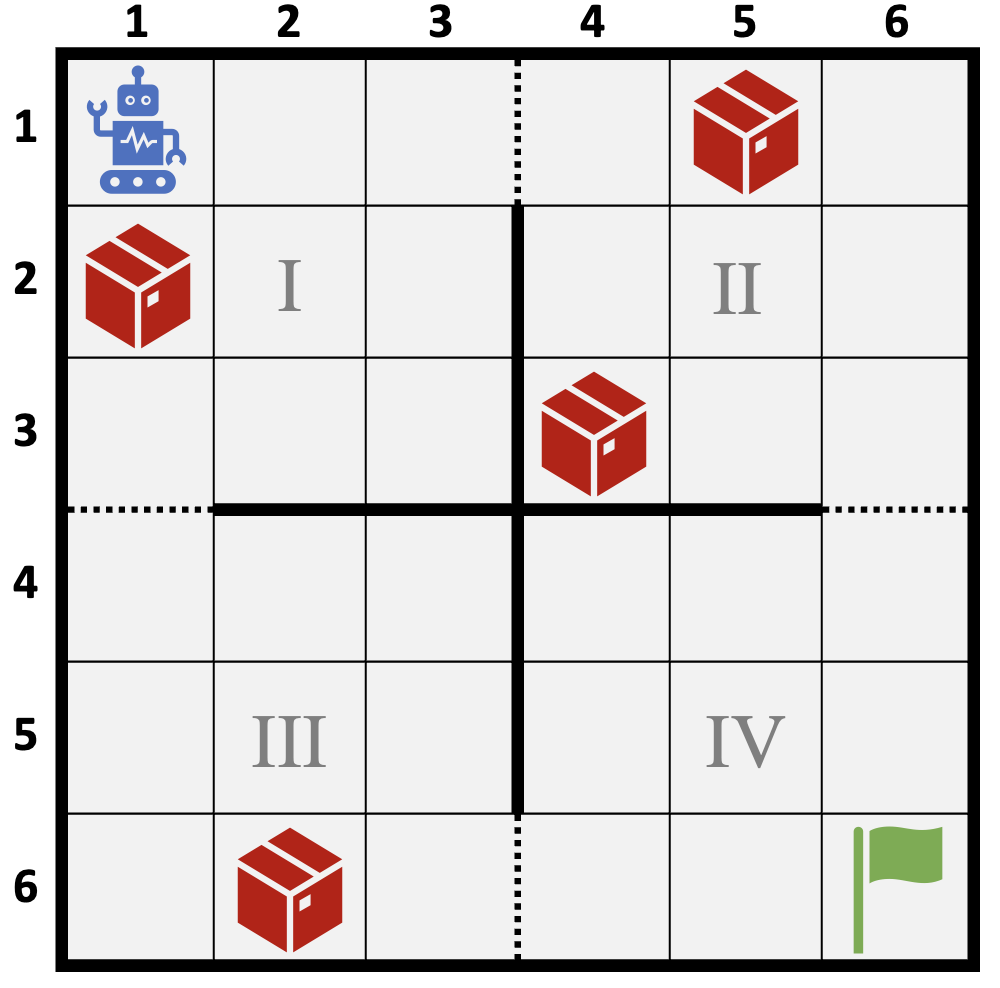}
    \caption{An example grid world environment where the robot aims to reach the flag while avoiding obstacles. The robot can move through the doors (dashed lines) between four rooms that are separated by walls (solid lines).}
    % \vspace{-10pt}
    \label{fig:grid}
\end{figure}

Consider an example of a robot navigating in the grid world environment shown in \figref{fig:grid}. 
Let $g_{ij}$ denote the grid location in row $i$ and column $j$. 
We model the environment as a POMDP with the state space $S = \{g_{ij}\}$ for $1 \le i,j \le 6$. 
The robot can take four actions to move \emph{east}, \emph{south}, \emph{west}, and \emph{north}, respectively. 
For each action, the robot moves to the target location with probability 0.8 or overshoots by one location with probability 0.2 due to slippery roads.
The robot starts in $g_{11}$ and aims to reach the flag in $g_{66}$ for which it will receive a reward of 1,000.
The step cost is 1 and the cost of colliding with an obstacle is 5. 
The robot's location during the trip is uncertain due to noisy sensors. 
Thus, we use POMDP belief states to represent the posterior probability distribution of the robot being in each grid given the history,
with the initial belief $b_0(g_{11})=1$.

Applying the POMCP algorithm to the above example yields one possible trajectory of the robot:
$g_{11} \xrightarrow{\mathit{east}} g_{13} \xrightarrow{\mathit{east}} g_{15} \xrightarrow{\mathit{east}} g_{16}
\xrightarrow{\mathit{south}} g_{26} \xrightarrow{\mathit{south}} g_{46} \xrightarrow{\mathit{south}} g_{66}$.
This trajectory exhibits unsafe behavior of the robot colliding into the obstacle in $g_{15}$.

In this work, we aim to tackle this problem by developing novel methods that can guarantee safety during the POMDP online planning. 

%-------------------
%-------------------

\section{Centralized Shielding} \label{sec:centralized} 

We consider safety requirements represented as almost-sure reach-avoid specifications (cf. \sectref{sec:reach-avoid}).
Given a POMDP and an almost-sure reach-avoid specification $\varphi$, we compute a (large) productive winning region $W_\varphi$ of the POMDP by applying the SAT-based iterative approach of~\cite{junges2021enforcing} incrementally to a fixpoint. 
We define a \emph{centralized shield} $\chi: b \to 2^A$ which, for any winning belief state $b$ of the POMDP, gives \emph{allowed} actions $\chi(b)$, which exclusively lead to belief support states within the winning region $W_\varphi$.
In practice, we do not compute such a shield explicitly. Instead, we only store the winning region $W_\varphi$, and decide whether to allow any encountered action on-the-fly by checking if $W_\varphi$ contains the resulting belief support states. 

We present two different ways of extending the POMCP algorithm with centralized shields for safe POMDP online planning, namely
\emph{prior pruning} and \emph{on-the-fly backtracking}, in \sectsectref{sec:prior}{sec:back}, respectively. 

%=============================================================================
\subsection{Prior Pruning} \label{sec:prior}

At each time step $t$, we find all actions disallowed by the shield $\chi(\beta(h_t))$ and prune the corresponding tree branches from the root node $\cT(h_t)$ before the POMCP algorithm iterations.
Specifically, for each action $a \in A$, we loop through every state $s \in \beta(h_t)$ and add successor states $s'$ generated by a black box simulator $(s',o,r) \sim \cG(s,a)$ to the set $\beta(h_tao)$. We check if these belief support updates are contained in the winning region $W_\varphi$.

% \blue{In the expansion phase of the root node (when root is a lead node), we find all actions disallowed by the shield 
% $\chi(\beta(h_t))$ and prune the corresponding tree branches. Specifically, for each action $a \in A$, we compute a set of updated belief support states $\beta(h_tao)$ and check if they are contained in the winning region $W_\varphi$ }

Following the motivating example, suppose that the history at time step $1$ is $h_1= \{\mathit{east}, g_{13}\}$, and the particle set is $\beta(h_1) = \{g_{12}, g_{13}\}$.
Consider an almost-sure reach-avoid specification $\varphi$ with $\reach = \{g_{66}\}$ and $\avoid = \{g_{15}, g_{21}, g_{34}, g_{62}\}$.
We find that the shield $\chi(\beta(h_1))$ disallows action \emph{east} because it may lead to a set of belief support states 
$\{g_{13},g_{14},g_{15}\}$ that are not contained in the winning region $W_\varphi$ (since $g_{15}$ is not a winning belief support state). 
We prune the tree branch of action \emph{east} at node $\cT(h_1)$. 
The unsafe robot trajectory in \sectref{sec:motiv} would be prevented.

\startpara{Correctness}
At each step $t$, any unsafe actions leading to non-winning belief support states are pruned. The agent selects the optimal action $a_t$ that is expected to yield winning beliefs, in an approximate sense, in accordance with the almost-sure reach-avoid specification $\varphi$. 
It is important to note that, as POMCP is a sampling-based algorithm, there is a risk that approximate belief states might overlook unsafe states, leading to false positives. However, with a sufficiently large particle set, the POMDP policy derived from centralized shielding with prior pruning, can offer safety guarantees.

%=============================================================================
\subsection{On-the-Fly Backtracking} \label{sec:back}

The prior pruning approach can only shield actions at the root node $\cT(h_t)$, without considering the safety of simulated paths,
which may cause the value of $V(h_ta)$ to be estimated based on unsafe simulations.  
To address this limitation, we propose the following on-the-fly backtracking procedure.

During the simulation phase of the POMCP algorithm, when an action $a$ is chosen (by the UCB rule or rollout) for history $h$
and a successor state $s'$ is generated by a black box simulator $(s',o,r) \sim \cG(s,a)$, we check if the updated particle set 
$\beta(hao) \cup \{s'\}$ is contained in the winning region $W_\varphi$. 
If the resulting particle set is not winning, we prune the tree branch starting from node $\cT(ha)$; 
that is, action $a$ would be shielded at node $\cT(h)$.

For example, suppose that state $s=g_{13}$ is sampled from the particle set $\beta(h_1) = \{g_{12}, g_{13}\}$ and action \emph{east} is chosen. 
Suppose that the black box simulator yields a successor state $s'=g_{15} \in \avoid$.
Thus, we would shield action \emph{east} at node $\cT(h_1)$ and prevent the robot colliding into the obstacle.

\startpara{Correctness}
At each step $t$, the selected optimal action $a_t$ has been encountered during the simulation and is allowed by a centralized shield via on-the-fly backtracking.  
Thus, the resulting POMDP policy is safe.
% (i.e., satisfies the almost-sure reach-avoid specification $\varphi$). 

% \startpara{Complexity}

% Pros: achieve higher expected return

% Cons: may require extra computation time since more simulations

\section{Factored Shielding} \label{sec:factored}

% Why need factored shielding? centralized shielding may not scale

The practical applicability of centralized shielding is limited by the computational effort required to obtain the winning region, which is correlated with the POMDP model size. 
To improve scalability, we develop a \emph{factored} shielding method, where we decompose a POMDP model into a set of smaller submodels (see~\sectref{sec:decompose}), compute a winning region for each submodel (see~\sectref{sec:fs-compute}), and integrate the set of obtained winning regions into the POMCP algorithm (see~\sectref{sec:fs-planing}). 
We show the correctness of the proposed method in \sectref{sec:fs-correct}.

%=============================================================================
\subsection{Decomposing a POMDP Model} \label{sec:decompose}

Given a POMDP model $\cM=(S, A, O, T, R, Z)$, we decompose it into a set of $N$ smaller POMDP models 
$\cM^i=(S^i, A^i, O^i, T^i, R^i, Z^i)$ for $1 \le i \le N$ 
based on the factorization of the state space such that 
$S = \bigcup_i S^i$.
We can leverage problem-specific knowledge to achieve an efficient decomposition scheme.

For example, we decompose the POMDP model $\cM$ of the motivating example (see~\sectref{sec:motiv}) into four submodels, corresponding to the four rooms shown in \figref{fig:grid}. 
The state space $S^1$ of submodel $\cM^1$ includes 9 grid locations covered by room I (i.e., $\{g_{ij}\}$ for $1 \le i,j \le 3$) and 4 outlet locations $\{g_{14}, g_{15}, g_{41}, g_{51}\}$ that can be reached by the robot when it takes an action in room I (e.g., moving east in $g_{13}$ or moving south in $g_{31}$). 
We include these outlet locations as absorbing states in $\cM^1$.
The action space is $A^1=A$.
The set $O^1 \subseteq O$ only captures relevant observations of submodel states $S^1$.
The transition function $T^1: S^1 \times A^1 \to \Dist(S^1)$, the reward function $R^1: S^1 \times A^1 \to \Rset$, and the observation function $Z^1: S^1 \times A^1 \to \Dist(O^1)$ are projections of the original POMDP model's transition function $T$, reward function $R$, and observation function $Z$ onto the submodel $\cM^1$, respectively. 
We define submodels $\cM^2$, $\cM^3$ and $\cM^4$ corresponding to rooms II, III and IV in a similar way.

%=============================================================================
\subsection{Computing Factored Winning Regions} \label{sec:fs-compute}

Given a set of factored POMDP models $\{\cM^i\}^N_{i=1}$ and an almost-sure reach-avoid specification $\varphi=\langle \reach, \avoid \rangle$, we compute a set of winning regions $\{W_\varphi^i\}^N_{i=1}$ to encode factored shields following \agref{alg:factored}.

First, for each model $\cM^i$, we determine the set of initial states $S^i_{\mathit{init}}$, reach states $S^i_{\mathit{reach}}$, and avoid states $S^i_{\mathit{avoid}}$.
Following the previous example, we define the initial states of $\cM^1$ as $S^1_{\mathit{init}} = \{g_{11}, g_{12}, g_{13}, g_{31}\}$, including the robot's initial location $g_{11}$ and three inlet locations that can be reached when the robot enters from an adjacent room. Note that we exclude the inlet location $g_{21} \in \avoid$ from $S^1_{\mathit{init}}$. 
We initialize the set of reach states as $S^i_{\mathit{reach}} = \reach \wedge S^i$ and the set of avoid states as $S^i_{\mathit{avoid}} = \avoid \wedge S^i$.
In this example, $\cM^4$ is the only model initialized with a non-empty reach set $S^4_{\mathit{reach}}=\{g_{66}\}$.

We compute a productive winning region $W_\varphi^4$ for the POMDP model $\cM^4$ by applying the SAT-based iterative approach in~\cite{junges2021enforcing} incrementally to a fixpoint. 
Since the robot may enter room IV directly from rooms II or III, we consider $\cM^2$ and $\cM^3$ as adjacent models of $\cM^4$ and add pairs $\langle 2,4 \rangle, \langle 3,4 \rangle$ to a queue. 

While the queue is non-empty, we remove the first element $\langle i,j \rangle$ of the queue and check if the reach set $S^i_{\mathit{reach}}$ should be updated based on the winning region $W_\varphi^j$. 
Suppose that $\langle 2,4 \rangle$ is removed from the queue and $S^2_{\mathit{reach}} = \emptyset$.
We find the winning region $W_\varphi^4$ containing the outlet states $\{g_{46}, g_{56}\}$ of model $\cM^2$ that the robot may encounter when moving from room II to room IV.
We update the reach set as $S^2_{\mathit{reach}}=\{g_{46}, g_{56}\}$ and compute an updated winning region $W_\varphi^2$.
Once a model's reach set and winning region are updated, we add all of its adjacent models to the queue. 
\agref{alg:factored} terminates when the queue is empty 
and returns the union of all factored winning regions $\bigcup_{i=1}^N W_\varphi^i$.

\begin{algorithm}[t]
\caption{Computing factored winning regions}\label{alg:factored}
\DontPrintSemicolon
\KwIn{A set of factored POMDP models $\{\cM^i\}^N_{i=1}$, an almost-sure reach-avoid specification $\varphi$}
\KwOut{A set of winning regions $\{W_\varphi^i\}^N_{i=1}$}
determine $S^i_{\mathit{init}}, S^i_{\mathit{reach}}, S^i_{\mathit{avoid}}$ of each model $\cM^i$ \;
$queue \gets [\ ]$ \;
\ForEach{model $\cM^j$ with a non-empty set $S^j_{\mathit{reach}}$} {
    compute winning region $W_\varphi^j$ \;
    \ForEach{adjacent model $\cM^k$ of $\cM^j$} {
        $queue.put(\langle k,j \rangle)$ \;
    }
}
\While{$queue$ is non-empty}{
    $\langle i,j \rangle = queue.get()$ \;
    check if $S^i_{\mathit{reach}}$ should be updated based on $W_\varphi^j$ \;
    \If{$S^i_{\mathit{reach}}$ has been updated} {
        compute the updated winning region $W_\varphi^i$ \;
        \ForEach{adjacent model $\cM^k$ of $\cM^i$} {
            $queue.put(\langle k,i \rangle)$ \;
        }
    }
}
\KwRet $\bigcup_{i=1}^N W_\varphi^i$
\end{algorithm}

%=============================================================================
\subsection{Shielding POMCP with Factored Winning Regions} \label{sec:fs-planing}

We integrate the POMCP algorithm with factored winning regions via prior pruning or on-the-fly backtracking similar to the centralized shielding method described in \sectref{sec:centralized}.

\startpara{Factored shielding with prior pruning}
At each step $t$, before the POMCP algorithm explore the tree with root node $\cT(h_t)$, we compute $\beta(h_tao)$ for each action $a \in A$ and prune any action that may lead to belief support states not contained in factored winning regions $\bigcup_{i=1}^N W_\varphi^i$.

\startpara{Factored shielding with on-the-fly backtracking}
During the POMCP simulation phase, we check if each successor state $s'$ generated by a black box simulator yields winning belief support updates contained in $\bigcup_{i=1}^N W_\varphi^i$.  
We shield any action that leads to non-winning simulated beliefs.

%=============================================================================
\subsection{Correctness} \label{sec:fs-correct}

\startpara{Lemma 1}
\emph{Given a POMDP model $\cM$ whose decomposition yields a set of factored models $\{\cM^i\}^N_{i=1}$, and an almost-sure reach-avoid specification $\varphi$, the output of \agref{alg:factored} forms a productive winning region of $\cM$ with respect to $\varphi$.
}

\startpara{\emph{Proof}}
Let $W:=\bigcup_{i=1}^N W_\varphi^i$ denote the output of \agref{alg:factored}.
We need to show that, from every belief support in $W$, there exists a finite path to stay within the region $W$ and reach some states in the set $\reach$ of $\varphi$, according to the definition of productive winning region (cf. \sectref{sec:reach-avoid}).

Consider a belief support $\Supp(b_0) \in W$ that belongs to the winning region $W_\varphi^i$ of model $\cM^i$.
Since $W_\varphi^i$ is productive towards the set $S^i_{\mathit{reach}}$ by construction, there exists a finite path from $\Supp(b_0)$ to $\Supp(b_k) \subseteq S^i_{\mathit{reach}}$ while staying within $W_\varphi^i$ along the path. 
If model $\cM^i$ was initialized with a non-empty set $S^i_{\mathit{reach}} = \reach \cap S^i$, then the path has reached the set $\reach$.
Otherwise, $S^i_{\mathit{reach}}$ was initialized as an empty set but updated with some winning belief support of adjacent model $\cM^j$. Thus, $\Supp(b_k) \in W_\varphi^j$ and there exists a finite path from $\Supp(b_k)$ to $\Supp(b_n) \subseteq S^j_{\mathit{reach}}$ while staying within $W_\varphi^j$ along the path.
We can prove by induction that the path would eventually reach some states in $\reach$ while staying within the region $W$. 

$\qed$

Based on the above lemma, we can follow the correctness argument of centralized shielding in \sectref{sec:centralized} to show that integrating the POMCP algorithm with factored shielding via prior pruning or on-the-fly backtracking yields safe POMDP policies that satisfy almost-sure reach-avoid specifications.  

\startpara{Remark}
We note that the union of factored winning regions computed by \agref{alg:factored} could be a subset of the winning region used in centralized shielding. 

\begin{figure}[t]
    \centering
    \includegraphics[width=0.6\columnwidth]{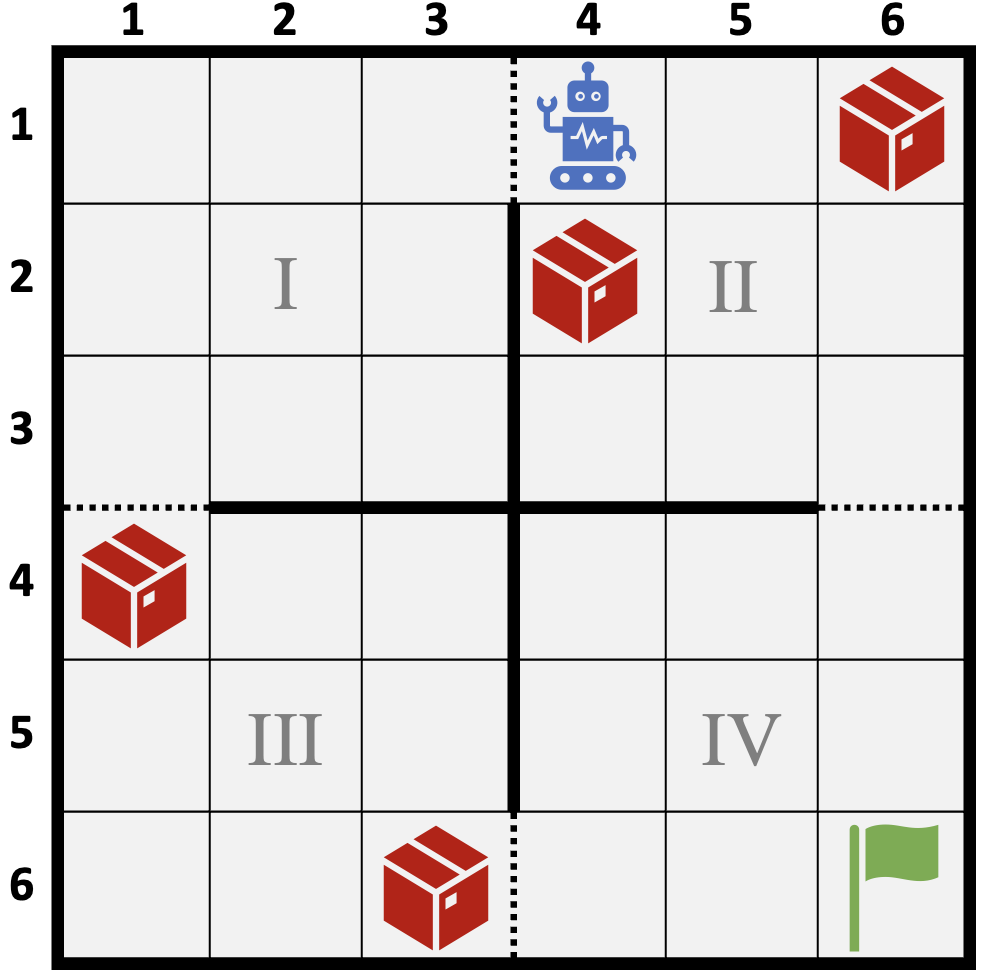}
    \caption{A grid world example for the comparison between centralized shielding and factor shielding.}
    % \vspace{-10pt}
    \label{fig:deadlock}
\end{figure}

Consider the grid world environment shown in \figref{fig:deadlock}, where the robot navigation model follows the one for the motivating example (see~\sectref{sec:motiv}).
When applying the centralized shielding method, we find that $g_{14}$ is a winning belief support state, since there exists a winning policy that yields the following trajectory:
$g_{14} \xrightarrow{\mathit{west}} g_{13} \xrightarrow{\mathit{east}} g_{15} \xrightarrow{\mathit{south}} g_{25} \xrightarrow{\mathit{east}} g_{26} \xrightarrow{\mathit{south}} g_{46} \xrightarrow{\mathit{south}} g_{66}$.

Now we apply \agref{alg:factored} for the factored shielding. 
We start by computing the winning region $W_\varphi^4$ of model $\cM^4$ corresponding to room IV. 
We update the reach set of model $\cM^2$ for the adjacent room II to $S^2_{\mathit{reach}}=\{g_{46}, g_{56}\}$ and compute the winning region $W_\varphi^2$.
We find that $g_{14} \not\in W_\varphi^2$, because there does not exist a safe path for the robot to reach $S^2_{\mathit{reach}}$ from $g_{14}$ while avoiding obstacles with probability one. 
The robot would collide with the obstacle in $g_{16}$ with probability 0.2 if it moved east, or collide with the obstacle in $g_{24}$ with probability 0.8 if it moved south.
Next, we update the reach set of model $\cM^1$ for room I to $S^1_{\mathit{reach}}=\{g_{15}\}$ and compute the winning region $W_\varphi^1$.
To make $g_{14}$ a winning belief support state, we would want $g_{12}$ and $g_{13}$ to be contained in $W_\varphi^1$ such that the robot can move west from $g_{14}$. 
Unfortunately, starting from $g_{12}$ or $g_{13}$, the robot is not guaranteed to reach $S^1_{\mathit{reach}}=\{g_{15}\}$ with probability one. 
Thus, $g_{14}$ is not contained the factored winning regions.

%-------------------
%-------------------

\section{Experiments} \label{sec:exp}

%-------------------
\begin{table*}[t]
\resizebox{\textwidth}{!}{
% \usepackage{multirow}
\begin{tabular}{cccccccccccccccccccc}
\toprule
\multicolumn{5}{c}{\textbf{Case Study}} &
  \multicolumn{3}{c}{\textbf{No Shield}} &
  \multicolumn{6}{c}{\textbf{Centralized Shield}} &
  \multicolumn{6}{c}{\textbf{Factored Shield}} \\ \midrule
\multicolumn{1}{l}{} &
  \multicolumn{1}{l}{} &
  \multicolumn{1}{l}{} &
  \multicolumn{1}{l}{} &
  \multicolumn{1}{l}{} &
  \multicolumn{1}{l}{} &
  \multicolumn{1}{l}{} &
  \multicolumn{1}{l}{} &
  \multicolumn{3}{c}{\textbf{Prior}} &
  \multicolumn{3}{c}{\textbf{On-the-fly}} &
  \multicolumn{3}{c}{\textbf{Prior}} &
  \multicolumn{3}{c}{\textbf{On-the-fly}} \\ \midrule
Domain &
  \multicolumn{1}{c}{Para.} &
  \multicolumn{1}{c}{$|S|$} &
  \multicolumn{1}{c}{$|O|$} &
  \multicolumn{1}{c|}{$|T|$} &
  \multicolumn{1}{c}{Time(s)} &
  \multicolumn{1}{c}{Return} &
  \multicolumn{1}{c|}{Unsafe} &
  \multicolumn{1}{c}{Time(s)} &
  \multicolumn{1}{c}{Return} &
  \multicolumn{1}{c|}{Unsafe} &
  \multicolumn{1}{c}{Time(s)} &
  \multicolumn{1}{c}{Return} &
  \multicolumn{1}{c|}{Unsafe} &
  \multicolumn{1}{c}{Time(s)} &
  \multicolumn{1}{c}{Return} &
  \multicolumn{1}{c|}{Unsafe} &
  \multicolumn{1}{c}{Time(s)} &
  \multicolumn{1}{c}{Return} &
  \multicolumn{1}{c}{Unsafe} \\ \midrule
\multirow{3}{*}{\begin{tabular}[c]{@{}c@{}}Obstacle\\ (N)\end{tabular}} &
  6 &
  37 &
  20 &
  \multicolumn{1}{c|}{204} &
  0.09 &
  978.9 &
  \multicolumn{1}{c|}{1.0} &
  0.12 &
  929.9 &
  \multicolumn{1}{c|}{0} &
  0.20 &
  968.1 &
  \multicolumn{1}{c|}{0} &
  0.12 &
  921.7 &
  \multicolumn{1}{c|}{0} &
  0.16 &
  968.7 &
  0 \\
 &
  8 &
  65 &
  20 &
  \multicolumn{1}{c|}{396} &
  0.14 &
  972.0 &
  \multicolumn{1}{c|}{2.0} &
  0.16 &
  977.1 &
  \multicolumn{1}{c|}{0} &
  0.19 &
  980.1 &
  \multicolumn{1}{c|}{0} &
  0.16 &
  962.9 &
  \multicolumn{1}{c|}{0} &
  0.21 &
  979.8 &
  0 \\
 &
  9 &
  82 &
  39 &
  \multicolumn{1}{c|}{464} &
  0.13 &
  974.0 &
  \multicolumn{1}{c|}{1.9} &
  0.18 &
  944.0 &
  \multicolumn{1}{c|}{0} &
  0.32 &
  962.0 &
  \multicolumn{1}{c|}{0} &
  0.18 &
  939.3 &
  \multicolumn{1}{c|}{0} &
  0.28 &
  964.0 &
  0 \\ \midrule
\multirow{3}{*}{\begin{tabular}[c]{@{}c@{}}Refuel \\ (N, E)\end{tabular}} &
  6, 8 &
  272 &
  74 &
  \multicolumn{1}{c|}{1,081} &
  0.84 &
  861.4 &
  \multicolumn{1}{c|}{20.4} &
  0.14 &
  795.9 &
  \multicolumn{1}{c|}{0} &
  0.53 &
  939.9 &
  \multicolumn{1}{c|}{0} &
  0.14 &
  912.8 &
  \multicolumn{1}{c|}{0} &
  0.53 &
  917.8 &
  0 \\
 &
  9, 6 &
  470 &
  151 &
  \multicolumn{1}{c|}{1,848} &
  1.19 &
  741.9 &
  \multicolumn{1}{c|}{40.0} &
  1.36 &
  -261.6 &
  \multicolumn{1}{c|}{0} &
  0.81 &
  904.5 &
  \multicolumn{1}{c|}{0} &
  1.41 &
  -259.6 &
  \multicolumn{1}{c|}{0} &
  0.81 &
  760.8 &
  0 \\
 &
  12, 8 &
  1,081 &
  180 &
  \multicolumn{1}{c|}{5,003} &
  1.54 &
  -199.0 &
  \multicolumn{1}{c|}{192.1} &
  - &
  - &
  \multicolumn{1}{c|}{-} &
  - &
  - &
  \multicolumn{1}{c|}{-} &
  1.15 &
  -248.4 &
  \multicolumn{1}{c|}{0} &
  0.62 &
  933.6 &
  0 \\ \midrule
\multirow{3}{*}{\begin{tabular}[c]{@{}c@{}}Rocks \\ (N, R)\end{tabular}} &
  6, 3 &
  4,157 &
  596 &
  \multicolumn{1}{c|}{4.3e\textbf{4}} &
  0.16 &
  1,001.1 &
  \multicolumn{1}{c|}{0.7} &
  0.33 &
  492.6 &
  \multicolumn{1}{c|}{0} &
  0.36 &
  1,020.9 &
  \multicolumn{1}{c|}{0} &
  0.32 &
  840.6 &
  \multicolumn{1}{c|}{0} &
  0.32 &
  1,062.3 &
  0 \\
 &
  8, 4 & 
  3.7e\textbf{4} &
  2,036 &
  \multicolumn{1}{c|}{4.7e\textbf{5}} &
  0.54 &
  1,013.3 &
  \multicolumn{1}{c|}{0.5} &
  - &
  - &
  \multicolumn{1}{c|}{-} &
  - &
  - &
  \multicolumn{1}{c|}{-} &
  0.99 &
  406.6 &
  \multicolumn{1}{c|}{0} &
  0.73 &
  1,091.5 &
  0 \\
 &
  9, 6 &
  1.2e\textbf{6} &
  3.3e\textbf{4} &
  \multicolumn{1}{c|}{1.8e\textbf{7}} &
  0.87 &
  1,008.0 &
  \multicolumn{1}{c|}{1.9} &
  - &
  - &
  \multicolumn{1}{c|}{-} &
  - &
  - &
  \multicolumn{1}{c|}{-} &
  1.33 &
  -119.0 &
  \multicolumn{1}{c|}{0} &
  1.40 &
  540.4 &
  0 \\ \bottomrule
\end{tabular}

\label{tab:exp}
}
\end{table*}
%-------------------

We built a prototype implementation of our shielding-enabled POMCP techniques as an extension of the PRISM model checking tool~\cite{KNP11}, with shields generated by the implementation from ~\cite{junges2021enforcing}.  We then evaluated it on a set of benchmark POMDP domains adapted from~\cite{junges2021enforcing}.

\begin{itemize}
    \item \textbf{Obstacle:} A robot navigates in $N \times N$ grid world environments, aiming to reach certain target locations while avoiding static obstacles. The reward function employed is identical to the one presented in the motivating example (cf. \sectref{sec:motiv}). 
    \item \textbf{Refuel:} A robot navigates in $N \times N$ grid world environments and consumes energy at every movement. The robot may recharge at refuel stations to the full battery capacity $E$. Noisy sensors introduce uncertainty about the robot's location and battery level. The robot's goal is to reach destinations while avoiding obstacles or running out of the battery. The cost structure is: moving incurs a cost of 1, idling also costs 1 due to the lack of energy to move, and refueling to full energy is set at a cost of 3 to discourage unnecessary refueling. The reward for reaching the target is set at 1000.
    \item \textbf{Rocksample:} A robot navigates in $N \times N$ grid world environments with $R$ rocks that are either valuable or dangerous to collect. To find out the quality of a rock with certainty, the robot has to sample it from an adjacent grid. The robot aims to reach target locations while avoiding sampling any dangerous rock.  The cost structure assigns a value of 1 to each action, whether moving, rock-sensing, or rock-sampling, and imposes a higher cost of 20 for encountering a bad rock. Successfully reaching the target yields a reward of 1000.
\end{itemize}

We use the following hyper-parameters for the POMCP algorithm: 
the number of simulations in each step's online planning is 40,000;
the simulation depth is 200;
and the number of particles sampled from the initial state distribution is 10,000.
We set the time-out for computing a winning region to be one hour. 
All experiments were run on a CentOS-7 machine with a 64GB Java memory limit.

\tabref{tab:exp} shows the experimental results.
For each POMDP model, we report the model parameters, the number of states $|S|$, observations $|O|$, and transitions $|T|$.
We compare the performance of the baseline (i.e., POMCP without shielding) and four variants of the proposed shielding methods in terms of the following metrics: search time per planning step, expected return, and occurrences of unsafe states. The results shown in \tabref{tab:exp} are the average over 10 runs of each method.  
We draw the following key insights from the results. 

\emph{The proposed shielding methods can guarantee safety but the baseline does not.}
Across all models, the POMCP algorithm without shielding yields non-zero occurrences of unsafe states, while all four variants of the proposed shielding methods avoid unsafe states completely. 

\emph{The proposed shielding methods have comparable search time per planning step with the baseline.}
This means that imposing shielding adds negligible overhead for the online planning. 
In some cases (e.g., Refuel(6,8)), shielding methods yield faster search than the baseline thanks to the pruning of tree branches. 
It only takes a few seconds to pre-compute winning regions for the shielding in most cases (except those time-out cases mentioned below).

\emph{Factored shielding has better scalability than centralized shielding.}
For some cases including Refuel(12,8), Rocksample(8,4), and Rocksample(9,6), centralized shielding fails to compute (fixpoint) winning regions before the time-out. 
By contrast, factored shielding scales up to large POMDP models with millions of states, since it only requires us to compute a set of smaller factored winning regions. 

\emph{On-the-fly backtracking generally yields higher expected return than prior pruning.}
Intuitively, on-the-fly backtracking chooses optimal actions based on node values estimated from safe simulated paths, while prior pruning only considers safety at the root node level and may choose locally optimal actions that cost more in the long run.

\section{Conclusion} \label{sec:conclu} 
In this work, we developed four distinct shielding methods, differing in how the shields are computed and integrated with the POMCP algorithm, for safe POMDP online planning with respect to almost-sure reach-avoid specifications. 
Experimental results on a set of benchmark domains demonstrate that the proposed shielding methods successfully guarantee safety (unlike the baseline POMCP without shielding), with negligible impact on the runtime for online planning.
In particular, factored shielding methods can scale up to solve large POMDP models with millions of states.

There are several directions to explore for possible future work. 
First, we will evaluate the proposed methods on a wider range of POMDP domains, beyond those benchmark domains considered in our experiments. 
Second, we will explore a principled way to achieve efficient POMDP model decomposition schemes for factored shielding. 
Finally, we would like to apply the proposed methods to robotic tasks in real-world scenarios (e.g., autonomous driving).

\section*{Acknowledgments}
This work was supported in part by U.S. National Science Foundation under grant CCF-1942836, 
U.S. Office of Naval Research under grant N00014-18-1-2829, 
U.S. Air Force Office of Scientific Research under grant FA9550-21-1-0164
and the ERC under the European Union’s Horizon 2020 research and innovation programme (FUN2MODEL, grant agreement No. 834115).
Any opinions, findings, and conclusions or recommendations expressed in this material are those of the author(s) and do not necessarily reflect the views of the grant sponsors.

%=============================================================================
\balance
\bibliographystyle{IEEEtran}
\bibliography{references}

\end{document}